 \documentclass[final,5p,times,authoryear]{elsarticle}

\usepackage{amsmath}
\usepackage{xspace}
\usepackage{todonotes}

\usepackage{siunitx}

\usepackage[english]{babel}

\usepackage[utf8]{inputenc}

\usepackage{caption}
\usepackage{tabularx} 

\usepackage{booktabs}

\graphicspath{{./images/}}

\usepackage{enumitem}

\usepackage{graphicx}
\usepackage{subcaption}

\usepackage[onehalfspacing]{setspace}

\def\BibTeX{{\rm B\kern-.05em{\sc i\kern-.025em b}\kern-.08em
    T\kern-.1667em\lower.7ex\hbox{E}\kern-.125emX}}
\usepackage{natbib}
\setcitestyle{numbers}

\graphicspath{{Images/}}

\usepackage{rotating}
\usepackage{pdflscape,kantlipsum}

\usepackage[T1]{fontenc}

\usepackage{makecell}
\usepackage{ifoddpage}

\usepackage{xcolor}

\usepackage{amsfonts}

\usepackage[bookmarks,bookmarksnumbered]{hyperref}
\hypersetup{colorlinks = true,linkcolor = blue,anchorcolor =red,citecolor = blue,filecolor = red,urlcolor = red,
            pdfauthor=author}
\usepackage{xurl}

\usepackage{microtype}

\usepackage{amssymb}
\usepackage{amsthm}
\usepackage{multirow}

\usepackage{amssymb}

\begin{document}

\begin{frontmatter}



\title{SynBench: A Synthetic Benchmark for Non-rigid 3D Point Cloud Registration}


\author[1]{Sara Monji-Azad\corref{cor1}} 
    \author[2]{Marvin Kinz}
    \author[6,7]{Claudia Scherl}
    \author[6]{David M\"annle}
    \author[1,3,4,5]{J\"urgen Hesser}
    \author[1]{Nikolas L\"ow}
    
    \cortext[cor1]{Corresponding author: Tel.: +496213838186; Email address: sara.monjiazad@medma.uni-heidelberg.de\\}
    
    \cortext[cor2]{Other email addresses: nikolas.loew@medma.uni-heidelberg.de;\\ juergen.hesser@medma.uni-heidelberg.de;mkinz@bwh.harvard.edu; \\Claudia.Scherl@umm.de; David.Maennle@medma.uni-heidelberg.de}

    \address[1]{Mannheim Institute for Intelligent Systems in Medicine (MIISM), Medical Faculty Mannheim, Heidelberg University, Mannheim 68167, Germany}
    \address [2]{Department of Radiation Oncology, Brigham and Women's Hospital, Dana-Farber Cancer Institute, Harvard Medical School, Boston, MA, USA}
    \address[3]{Interdisciplinary Center for Scientific Computing (IWR), Heidelberg University, Heidelberg, Germany}
    \address[4]{Central Institute for Computer Engineering (ZITI), Heidelberg University, Heidelberg, Germany}
    \address[5]{CZS Heidelberg Center for Model-Based AI, Heidelberg University, Mannheim, Germany}
    \address[6]{Department of Otorhinolaryngology, Head and Neck Surgery, University Hospital Mannheim, Medical Faculty Mannheim, Heidelberg University, Mannheim, Germany}
    \address[7]{AI Health Innovation Cluster, Heidelberg-Mannheim Health and Life Science Alliance, Heidelberg University, Heidelberg, Germany}

\begin{abstract}
Non-rigid point cloud registration is a crucial task in computer vision. Evaluating a non-rigid point cloud registration method requires a dataset with challenges such as large deformation levels, noise, outliers, and incompleteness. Despite the existence of several datasets for deformable point cloud registration, the absence of a comprehensive benchmark with all challenges makes it difficult to achieve fair evaluations among different methods. This paper introduces SynBench, a new non-rigid point cloud registration dataset created using SimTool—a toolset for soft body simulation in Flex and Unreal Engine. SynBench provides the ground truth of corresponding points between two point sets and encompasses key registration challenges, including varying levels of deformation, noise, outliers, and incompleteness. To the best of the authors' knowledge, compared to existing datasets, SynBench possesses three particular characteristics: 1) it is the first benchmark that provides various challenges for non-rigid point cloud registration, 2) SynBench encompasses challenges of varying difficulty levels, 3) It includes ground truth corresponding points both before and after deformation. The authors believe that SynBench makes it possible for future non-rigid point cloud registration methods to present a fair comparison of their achievements. The SynBench is publicly available under \url{https://doi.org/10.11588/data/R9IKCF           
}.

\end{abstract}




\end{frontmatter}



\section{Introduction} \label{Introduction}
In recent years, 3D point cloud registration has gained substantial interest in various research fields such as 3D reconstruction \citep{takimoto20163d}, augmented reality \citep{mahmood20193d}, tracking \citep{wang20193d} and generation of free-viewpoint videos \citep{zhang2021representation}, to mention a few. \par

The primary objective of the point cloud registration problem is to find the transformation between two or more point sets, typically referred to as source and target data. This challenge can be addressed through either learning-based or non-learning-based approaches \citep{monji2023review}. Non-learning approaches primarily rely on iterative optimization solutions \citep{besl1992method}, while learning-based approaches involve training models to learn the transformation of the point clouds \citep{Feng.2019}. In addition to determining the transformation between source and target point sets, another key goal in this field is to overcome various existing challenges, which can be categorized into different types, as presented below \citep{monji2023review}.

\textbf{\textit{Different deformation levels:}} Object deformation refers to changes in the size, shape, or volume of a soft object. In applications such as medical surgery, soft tissues often undergo significant deformation. Therefore, it's crucial to assess the robustness of registration methods to various deformation levels, particularly when handling substantial deformations. \par

\textbf{\textit{Different levels of noise and outliers:}} Robustness to varying levels of noise and outlier ratios represents another significant challenge. Given that some approaches rely on identifying corresponding points, both noise and outliers can significantly impact the accuracy of the registration method. \par

\textbf{\textit{Incompleteness and partial overlapping:}} Aligning two partial point clouds poses another challenging task. Data incompleteness \citep{wang2020unsupervised} and partial overlapping of point sets \citep{zhou2013dense} can be particularly challenging for registration methods, especially when dealing with applications that involve partial observations. \par

To evaluate both learning-based and non-learning approaches, various datasets have been published. Nevertheless, the field still faces a shortage of a benchmark that encompasses most of the challenges mentioned simultaneously. A benchmark, in this context, refers to a dataset that can be employed for assessing different methods. Such a benchmark enables fair comparisons between various approaches. However, creating a benchmark can be more complex in certain domains, particularly in areas like the medical field and surgical applications.\par

In a prior study by the authors, we introduced a toolset named SimTool \citep{monji2023simtool}, which is a toolbox designed for simulating soft body deformation and generating deformable point clouds. In this paper, SimTool is employed to create a benchmark for non-rigid point cloud registration, known as SynBench. The proposed SynBench serves to evaluate point cloud registration methods. Subsequently, we describe the development process of the generated dataset, making it practical to modify and adapt for various applications. \par

To make the novelty of this article clear, the main contributions are summarized below:
\begin{itemize}
\item \textit{\textbf{A benchmark of definable objects:}}
although the generated dataset is generated using a simulation of soft body deformation, it is possible to generalize the dataset in further applications. Unlike datasets with predefined objects, such as animals or human bodies, the proposed dataset can be used in any application of non-rigid objects to train machine learning models.

\item \textit{\textbf{Challenges and robustness:}}
The generated benchmark encompasses various challenges in point cloud registration, including diverse deformation levels, varying levels of noise, outlier ratios, and data incompleteness. This comprehensive approach allows us to assess the robustness of methods to these challenges effectively.

\item \textit{\textbf{Ground truth of corresponding points:}}
The dataset provides ground truth corresponding points for both pre-deformation and post-deformation objects and slices. This valuable information can be used not only to evaluate registration method accuracy but also for other 3D point cloud applications.
\end{itemize}

The rest of this paper is organized as follows: Section \ref{Related Work} provides an overview of the related studies on available datasets. The structure of the SynBench, as well as the benchmark characteristics, are provided in Section \ref{SynBench}. In Section \ref{Evaluations}, an evaluation of SynBench and its features are presented. Finally, different aspects of the generated dataset are discussed in Section \ref{Discussion}, and a conclusion is presented in Section \ref{Conclusion}.

\section{Related work} \label{Related Work}
The accuracy of non-rigid point cloud registration methods is reported using different available datasets. In other words, due to the fact that most of the challenges are not provided in existing datasets, the methods customize them to show how robust they are to the various challenges. The available datasets are divided into two main categories, namely real-world datasets and synthetic ones \citep{monji2023review}. In the following, the most frequently used datasets of non-rigid point cloud registration methods are discussed. Furthermore, a comparison between the SynBench dataset and some available benchmarks for non-rigid point cloud registration is presented in Table \ref{table:DatasetTable}.\par

\begin{table*}[!ht]
\caption{Comparison between SynBench dataset and some available benchmarks for non-rigid point cloud registration}
\label{table:DatasetTable} 
\centering
\small
\begin{tabular}{p{2.75cm} p{2cm} p{2cm} c c c c c } \hline
\hline 
Dataset Name (Ref)& Synthetic/Real & {\centering Matching\\correspondences}&Deformation levels&Noise&Outliers&Incompleteness&Overlapping\\ [0.5ex] 
\hline 

3DLoMatch \citep{ShengyuHuang.} &Real&\centering\checkmark&-&-&-&-&-\\ [0.25ex]
\hline 

Faust \citep{bogo2014faust} &Real&\centering\checkmark&-&-&-&-&-\\ [0.25ex]
\hline 

SCAPE \citep{anguelov2005scape}&Synthetic&\centering\checkmark&-&-&-&-&-\\ [0.25ex]
\hline 

4DMatch/4DLoMatch \citep{li2022lepard}&Synthetic&\centering\checkmark&-&-&-&-&\checkmark\\ [0.25ex]
\hline 

SyBench (Ours)&Synthetic&\centering\checkmark&\checkmark&\checkmark&\checkmark&\checkmark&-\\ [0.25ex]
\hline 

\end{tabular}
\end{table*}


\textit{{\textbf{Real-world datasets.}}} Representing the objects and some knowledge about them to provide an understanding of the real scene is called the real-world dataset. Real-world datasets usually are captured using LiDAR cameras, navigating robots equipped with cameras, or any kind of depth cameras, to mention a few \citep{handa2016understanding}. \par

3DMatch \citep{zeng20173dmatch} is a popular real-world dataset, which provides two groups of benchmarks, namely a keypoint matching benchmark and a geometric registration one. This dataset provides depth and color information as well. 3DLoMatch \citep{ShengyuHuang.} is another real-world dataset for the registration task which includes 62 scenes with an overlap ratio between 10 to 30 percent. Faust \citep{bogo2014faust} is another dataset that represents real human meshes with ground-truth correspondences. It has 300 triangulated meshes of 10 various subjects, five males and five females. Each scan is in 30 different poses. 

\textit{{\textbf{Synthetic datasets.}}}
Synthetic datasets use data that is artificially generated by using computer algorithms. Several synthetic datasets are available, which are mentioned in the following. \par

Chui-Rangarajan dataset \citep{chui2000new} is commonly used in registration articles. Chui-Rangarajan contains Chinese characters and fish shapes. Another available dataset is SCAPE \citep{anguelov2005scape}, which has generated meshes from a person in different poses. It is a benchmark for the matching problem and non-rigid registration. It includes 71 registered meshes of a particular person in different poses, in which the full objects' correspondences are provided. Furthermore, 4DMatch/4DLoMatch \citep{li2022lepard} are two other benchmarks for registration and matching problems. It can be used both for rigid and deformable scenes. 4DMatch is a partial point cloud benchmark while its low overlap version is called 4DLoMatch. It includes 1761 randomly selected sequences from DeformingThings4D. 4DLoMatch has low overlap in comparison with 4DMatch.

As shown in Table \ref{table:DatasetTable}, although most of the available benchmarks for non-rigid point cloud registration provide matching correspondences, the most important challenges like different deformation levels, noise, outlier, and incompleteness are not presented. What makes our proposed dataset (SynBench) most remarkable is not only the matching corresponding points available but also the significant challenges in non-rigid registration. SynBench dataset makes it possible for various approaches to have a fair comparison of their approach advantages. 

\section{SynBench - A Synthetic Benchmark} \label{SynBench}

In this section, the benchmark called SynBench, which is generated by using SimTool, will be described. The main aim of this section is to explain the structure and characteristics of SynBench. Although SynBench is a dataset for the simulation of soft body deformation, it can be considered a definable object benchmark and can be used in further applications besides tissue modeling like training machine learning models. \par

The proposed dataset is generated based on 30 primitive objects which are received from SimTool outputs. These primitive objects will be used to generate different deformation levels. Furthermore, some challenges will be presented that can be utilized for various approaches. Different outlier ratios, various noise levels, and incompleteness can be mentioned as the presented challenges in the SynBench dataset. In addition, point correspondence ground truth will be presented. \par
Considering the fact that all challenges are generated under small to large deformation levels, the SynBench makes it possible for users to select their desired dataset based on the ability of their proposed method, and to show how robust to complex challenges their methods are. \par

The SynBench dataset consists of five main sets, namely "Data" which represents 30 primitive objects as well as "Deformation Level", "Incompleteness", "Noise", and "Outlier" categories, including 30, 5302, 26515, 21213 and 26500 object samples, respectively. Each object sample includes a pair of source and target point clouds, which means that the total number of files is double. The reason that the number of files for some challenges has increased is that each challenge is applied on all different deformation levels as well as considering the effective parameter changes. The effective parameters for each challenge will be discussed in the following.\par

\subsection{\textbf{Different deformation levels}}
To generate different deformation levels, two strategies are implemented. First, this process is simulated by varying the gravity using Flex in the UE4, as explained in \citep{monji2023simtool}. Therefore, 30 primitive objects are generated in this way. In the second step, a geometric transformation called thin-plate spline (TPS) \citep{chui2003new} is used to generate a dataset in different deformation levels, which will be presented in the following. To estimate the deformation level, some methods calculate the distance between the source point set and the deformed one. The Hausdorff distance is one of the quantitative evaluations, which is widely used in articles. In this paper, the Hausdorff distance is used to show the deformation level between every two point sets. The Hausdorff distance between two point sets \textit{($X$, $Y$)} is defined as \textit{HD\ ($X$, $Y$)} where \textit{d($Y$, $X$)} and \textit{d($X$, $Y$)} are symmetrically defined. Furthermore, $\parallel . \parallel$ demonstrates the Euclidean norm. The definition of Hausdorff distance is shown in Equation \ref{equ1}:

\begin{align} \label{equ1}
\begin{split}
&HD\ (X, Y)= \max\ [d(X, Y),\ d(Y, X)],\\
&d(X, Y) = max_{p_1\in X}\ d(p_1, Y),\\
&d(Y, X) = max_{p_2\in Y}\ d(p_2, X),\\
&(p1, Y) = min_{p_2\in Y}  \parallel p_1-p_2 \parallel,\\
&(p2, X) = min_{p_1\in X}  \parallel p_2-p_1 \parallel
\end{split}
\end{align}

\begin{figure}[h]
	\centering
        \centering
        \includegraphics[width=9cm]{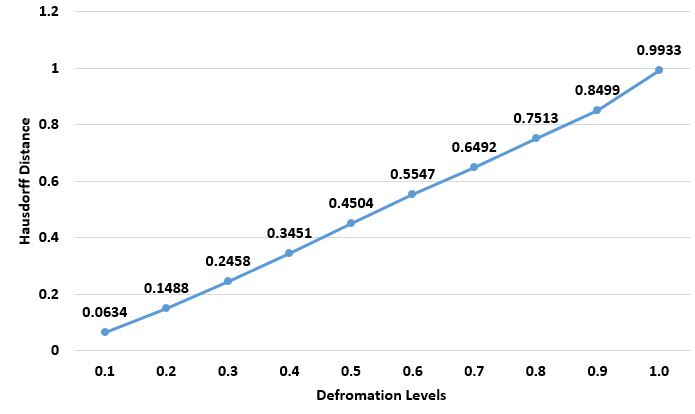}
        \caption{Hausdorff distance for different deformation levels in the generated dataset}
        \label{fig:HausDistanceLoss}
\end{figure}

\textbf{\textit{Using thin-plate spline}.}
Thin-plate spline (TPS) has a representation in terms of radial basis functions. TPS is a practical approach that can be decomposed into affine and non-affine components. Therefore, it can present a natural deformation field. In this aspect, the non-rigid geometric transformation in different deformation levels is simulated using TPS inspired by \citep{wang2019coherent}. To achieve this purpose, the control points are selected using a uniform distribution. Then, a Gaussian random shift is applied on each control point with \textit{zero-mean} and different standard deviations. As the next step, TPS is applied to deform point sets using mentioned initial values. The number of control points, the standard deviation, and the width of the Gaussian radial basis function can affect the deformation results. Small to large deformation levels for different initial shapes are shown in Figure \ref{fig:HausDistanceLoss}.  

\subsection{\textbf{Different levels of noise and outliers}} \label{Noiseoutlier}
Noise is one of the significant challenges in different applications. Traditionally, articles report how robust to noise the methods are. Due to the fact that the current dataset is generated synthetically, the initial point clouds are free of noise. Therefore, different amounts of noise can be added synthetically to the dataset. In this way, Gaussian noise is applied to point sets with zero mean and different standard deviations. Standard deviations between 0.01 and 0.04, are common in papers to show small to large amounts of noise, e. g., \citep{li2021pointnetlk}, \citep{aoki2019pointnetlk}, to mention a few. Therefore, the noisy dataset is generated in 4 categories, including 0.01, 0.02, 0.03, and 0.04 (std) noise per point set.\par


Outliers are another important challenge in various point cloud applications. An outlier is a data point that is significantly different from other observations. To generate the dataset with outliers, a Gaussian distribution with different means and standard deviations are used. Therefore, different amounts of points are randomly added to each point set. Considering the available studies, e. g., \citep{cheng2017automatic}, \citep{billings2015iterative}, the number of outliers can vary between 5\% to 50\%. Therefore, 5 categories are created with different amounts of outliers, including 5\% to 45\% outliers in steps of 10, per point set. \par

Nevertheless, it is worth mentioning that for various applications each method can select the desired amount of noise and outliers to show how robust the approach is. 

\subsection{\textbf{Incompleteness}}\label{Incompleteness}
Data incompleteness is another challenge for point cloud applications. To simulate this challenge, two different approaches are considered. One approach to simulate the data incompleteness challenge which was explained in \citep{monji2023simtool}. In the second approach, a ratio of points is deleted randomly. Therefore, 5 categories are generated in which 5\%, 10\%, 15\%, 20\%, and 25\% of points per point set are removed.\par

\subsection{\textbf{Ground truth of corresponding points}}
\begin{figure}[htp]
	\centering
        \centering
        \includegraphics[width=8cm]{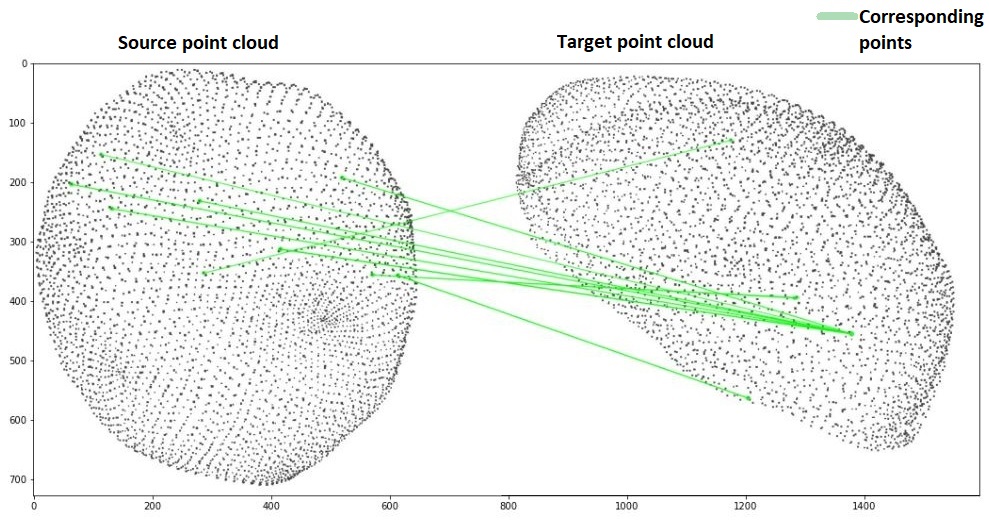}
        \caption{ground truth of corresponding points is available. To have a better visualization only some corresponding points are shown.}
        \label{fig:Correspondingpoints}
\end{figure}

For the dataset which is generated using SimTool, the ground truth of corresponding points is available for whole, undeformed objects to deformed objects and undeformed slices to deformed slices, excluding the sampled cut surfaces. As shown in Figure \ref{fig:Correspondingpoints}, the ground truth of the correspondences can be useful to evaluate different methods such as matching and registration approaches. Furthermore, for the generated SynBench, the corresponding points are available for primitive objects to deformed ones. Also, the coordinates of the outlier are presented in additional files.

\section{Evaluations} \label{Evaluations}
To evaluate the proposed dataset, two approaches are considered. First, dataset characteristics and features are provided including evaluation by area and broad-narrow. Second, an evaluation of a baseline method \citep{hansen2021deep} on different challenges of SynBench is presented. 

\subsection{\textbf{Dataset characteristics and features}}
In this section, the primitive data will be evaluated by different features for point cloud categorizations. Point clouds that are generated by SimTool are split up into sources and targets, and each point cloud is classified by two features. This information together with the camera positions are stored in an accompanying .csv file. The two features are \textit{Area} and an indicator for whether the points are broad clusters around their center or narrow lines, called \textit{Broad-Narrow} parameter. The first one is to handle the point clouds by their covered area, however, the second one can be used as an additional size parameter, because very narrow point clouds can cover a big area and the other way around.\par
\textbf{\textit{Evaluation by Area}.} 
The area $A_v$ for each partial point cloud is approximated by first determining the area $A_c$ of the complete surface and then multiplying it with the number of visible points $N_v$ over the number of total points $N_c$, as shown in Equation \ref{equ2}. 

\begin{equation} \label{equ2}
A_v=A_c\frac{N_v}{N_c}
\end{equation}

\begin{figure}[htp]
	\centering
        \centering
        \includegraphics[width=8cm]{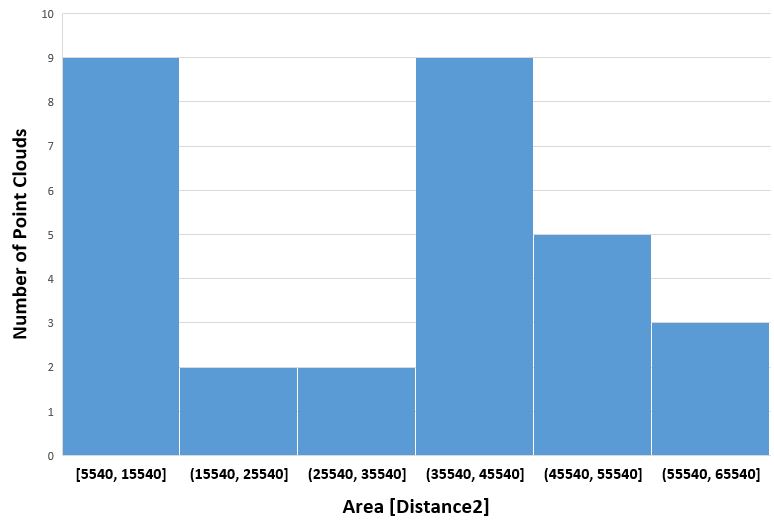}
        \caption{Histogram of the Area for primitive data of SynBench. The y axis displays the number of point clouds and the x-axis the area in simulation units with the dimension $distance^2$.}
        \label{fig:HistogramArea}
\end{figure}

A histogram of the areas for primitive data of SynBench is shown in Figure \ref{fig:HistogramArea}. The \textit{area} amount between 5540-15540 shows the smallest ones and the areas between 55540-65540 show the largest ones. 
One should also note, that the area alone does not tell a lot about what the point cloud will look like because it depends on the size of the cut. So if one has a very big cut even the small border captures could have a relatively big area compared to this sample graph and the other way around. This is why in addition the \textit{Broad-Narrow} parameter has been introduced, which will be explained in the next section.

\textbf{\textit{Evaluation by Broad or Narrow}.}
The \textit{Broad-Narrow} parameter \textit{bn} gives an approximation of whether the capture is more of a broad point cluster or a long and narrow sample, which for example happens when the side of the surface is captured. It is calculated by determining the mean distance of all points $\overrightarrow{x_i}$ from their center $\overrightarrow{x_m}$ and then dividing this by the square root of the area $A_v$ to make it independent and dimensionless. It is shown in Equation  \ref{equ3}, where $N_v$ is the number of points. This results in a small value for broad clusters and a higher value for narrow arrangements.

\begin{align}\label{equ3}
\begin{split}
\overrightarrow{x_m}=\frac{\sum_i{\overrightarrow{x_i}}}{N_v} \\
bn=\frac{1}{N_v\sqrt{A_v}}\sum_i{\lvert\overrightarrow{x_m}-\overrightarrow{x_i}\rvert}
\end{split}
\end{align}

\begin{figure}[h]
	\centering
        \centering
        \includegraphics[width=7.5cm]{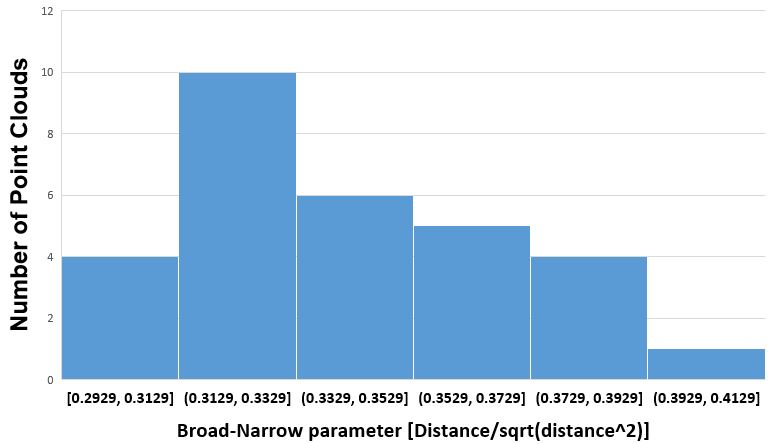}
        \caption{Histogram of the Broad-Narrow parameter. The y-axis displays the number of point clouds and the x-axis Broad-Narrow parameter, which is dimensionless [$\frac{distance}{\sqrt{distance^2}}$]}
        \label{fig:HistogramBroadNarrow}
\end{figure}

The histogram of the \textit{Broad-Narrow} parameter is shown in Figure \ref{fig:HistogramBroadNarrow}. Every point cloud with a value below one has at least some broad part, whereas a higher value is mostly narrow border captures.

\begin{figure}[htp]
	\centering
        \centering
        \includegraphics[width=8cm]{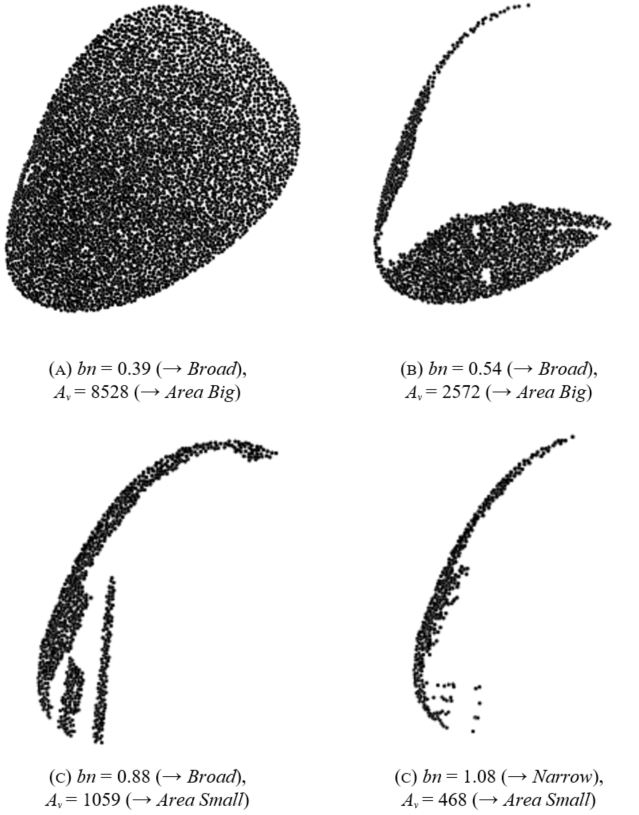}
        \caption{Point clouds with different \textit{Broad-Narrow bn} and \textit{Area $A_v$} values}
        \label{fig:ResultsParameters}
\end{figure}

Some outputs of point clouds with different \textit{Broad-Narrow bn} and \textit{Area $A_v$} values are shown in Figure \ref{fig:ResultsParameters}. This is how one can use those two parameters together to sort or categorize the point clouds to customize the requirements of registration.

\subsection{\textbf{Evaluation of Different Challenges}}
As discussed, SynBench is the first benchmark for non-rigid point cloud registration which includes various challenges. SynBench makes it possible for future methods to present a fair comparison of their achievement. To exhibit the usage of the proposed dataset (SynBench), a baseline approach for different presented challenges is run. 
A non-rigid point cloud registration method is proposed in \citep{hansen2021deep}. The method is a combination of using edge convolutions \citep{wang2019dynamic} to extract geometric features, and differentiable loopy belief propagation (LBP) networks \citep{monji2023review}. \par

\begin{figure*}[ht]
    \centering
    \includegraphics[width=17cm]{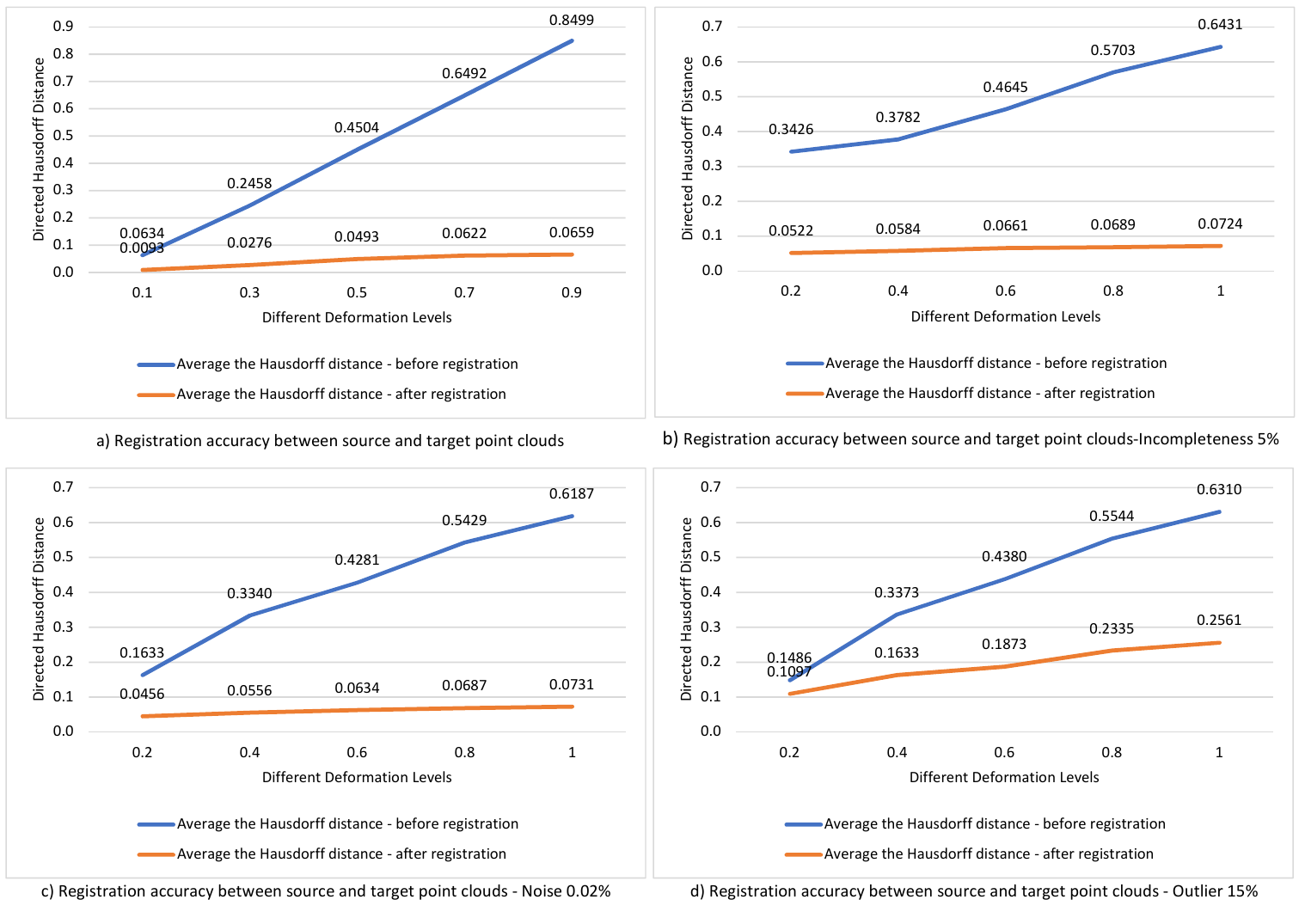}
    \caption{The robustness of the method \citep{hansen2021deep} to different challenges on SynBench dataset. The accuracy of the method in different deformation levels is shown in (a). The accuracy of the method when there is $5\%$ incompleteness in different deformation levels is shown in (b). The accuracy of the method to 0.02 noise levels and $15\%$ outliers ratios in different deformation levels are shown in sub-figures (c) and (d), respectively.}
    \label{fig:ChallengesEval}
\end{figure*}

As discussed in Section \ref{SynBench}, SynBench includes various challenges, namely different deformation levels, different ratios of incompleteness, and different noise and outlier levels. In figure \ref{fig:ChallengesEval}, the accuracy of method \citep{hansen2021deep} and its robustness to different challenges are shown. To study the result, all challenges are studied at different deformation levels. To this end, for each challenge, one specific ratio is considered. As shown in Figure \ref{fig:ChallengesEval}, the registration results of the method \citep{hansen2021deep} on SynBench are quite acceptable. The results for the incompleteness challenge are reported on $5\%$ category. The way of generating different categories with various incompleteness ratios was described in Section \ref{Incompleteness}. Furthermore, the accuracy of the method on SynBench when the data includes outliers and noise is shown in Figure \ref{fig:ChallengesEval}. The method accuracy in different deformation levels, when 0.02 noise level is added to the point cloud, is shown in Figure \ref{fig:ChallengesEval}. C. Finally, in Figure \ref{fig:ChallengesEval}. D, it is demonstrated that if the $15\%$ outlier adds to the data, the method is still robust to different deformation levels. An explanation of generating different categories of noise and outliers was presented in \ref{Noiseoutlier}.

\section{Discussion} \label{Discussion}
While there are several datasets used for evaluating point cloud registration methods, the absence of a comprehensive benchmark covering all aspects of the existing challenges is a notable issue. Therefore, the main objectives of this article are to create a dataset suitable for non-rigid point cloud registration and to incorporate most of the prevailing challenges. In this regard, the proposed SimTool \citep{monji2023simtool} allows for customizable steps in deformable object generation, and the provided SynBench dataset facilitates fair comparisons among available approaches. SynBench not only addresses the mentioned challenges but also introduces varying levels of difficulty for each of them. This diversity in challenge levels enables us to assess method robustness and determine their limitations.\par

Another goal of the developed dataset is to simulate natural deformation by applying physical properties using Flex in UE4. While deforming an object using physical properties is generally correct, it's essential to clarify the interpretation of these properties. Although Flex can be used for visually pleasing soft body deformation, it's important to note that the settings of a position-based dynamics (PBD) solver cannot be directly equated with physical properties. In this context, deformation is achieved synthetically by adjusting physics properties, such as gravity, as demonstrated in \citep{monji2023simtool}. Furthermore, when considering object deformation observed using Flex in UE4, it's evident that synthetic cutting of an object can alter deformation patterns. This cutting process is simulated as a slicing operation, as shown in \citep{monji2023simtool}.\par

Furthermore, the generated dataset simulates incompleteness, a significant challenge in point cloud applications. This challenge can occur in applications such as multi-view registration and 3D multi-view reconstruction, among others. To simulate point cloud incompleteness, different camera positions are used to capture objects from various points of view. It's worth noting that, from some perspectives, the captured areas may be very small and narrow, such as the edges of cuts shown in Figure \ref{fig:ResultsParameters} (c). Recognizing that these narrow areas are unsuitable for registration, specific thresholds are defined to facilitate their easy filtration. SimTool takes this into account, allowing users to define and generate their datasets for multi-view purposes.\par

Additionally, while the primary goal of the proposed dataset in this paper is to introduce a benchmark for point cloud registration, it's worth noting that this dataset can also find utility in various other applications. Given that it includes different points of view of the same object as well as its complete version, the dataset can be employed for additional purposes. Examples include 3D object reconstruction from various viewpoints and partial surface registration. \par

\section{Conclusion} \label{Conclusion}
Non-rigid point cloud registration is a crucial topic in computer vision. In recent years, various datasets have been created to assess registration methods. In this paper, we introduce a novel synthetic dataset called SynBench for non-rigid point cloud registration. SynBench encompasses essential challenges with varying levels of difficulty in point cloud registration, which have not been previously featured in other publications. While SimTool was originally designed to simulate soft body deformation, it has evolved to allow dataset customization for broader applications. Consequently, SynBench is a dataset that not only represents soft body deformation but also lacks predefined shapes like human bodies or animals. This versatility positions SynBench as a dataset suitable for evaluating registration methods across diverse applications.

\bibliography{elsarticle-template-harv}

\end{document}